\pdfoutput=1

\documentclass[11pt]{article}

\PassOptionsToPackage{table}{xcolor}
\usepackage[preprint]{acl}
\usepackage{enumitem}
\usepackage{times}
\usepackage{latexsym}
\usepackage{booktabs}
\usepackage{xcolor}
\usepackage{tcolorbox}
\usepackage{arydshln}

\usepackage[utf8]{inputenc}

\usepackage{microtype}

\usepackage{inconsolata}          
\usepackage{pifont}             
\usepackage{graphicx}           
\usepackage{multirow} 
\usepackage{algorithm}
\usepackage{algpseudocode}
\usepackage{tikz}
\usepackage{colortbl}
\usepackage{xcolor}
\usepackage{tcolorbox}
\tcbuselibrary{most}
\usepackage{graphicx}
\newcommand{\datasetname}{\textsc{TrustSim}}
\usepackage{pifont}
\usepackage{amsmath}
\newcommand{\Checkmark}{\ding{51}}

\newcommand{\largedot}{\scalebox{2.0}{$\bullet$}}

\usepackage{todonotes}
\usepackage{xcolor}

\definecolor{c1}{HTML}{F8EEEC}
\definecolor{c2}{HTML}{F0EDDE}

\title{Social Science Meets LLMs: How Reliable Are Large Language Models in Social Simulations?}

\author{
\textbf{Yue Huang}$^{1}$\footnotemark[1], ~~\textbf{Zhengqing Yuan}$^{1}$\footnotemark[1], ~~\textbf{Yujun Zhou}$^{1}$\footnotemark[1],  \textbf{Kehan Guo}$^{1}$\\ \textbf{Xiangqi Wang}$^{1}$, \textbf{Haomin Zhuang}$^{1}$, \textbf{Weixiang Sun}$^{1}$, \textbf{Lichao Sun}$^{2}$ \\ \textbf{Jindong Wang}$^{3}$, \textbf{Yanfang Ye}$^{1}$, \textbf{Xiangliang Zhang}$^{1}$\footnotemark[2] \\
\\
$^{1}$University of Notre Dame~ \\ $^{2}$Lehigh University~ $^{3}$William \& Mary\\
\texttt{\{yhuang37,zyuan2,yzhou25,xzhang33\}@nd.edu} 
}

\begin{document}
\maketitle
\renewcommand{\thefootnote}{\fnsymbol{footnote}}
\footnotetext[1]{Equal contribution.}
\footnotetext[2]{Corresponding author.}

\begin{abstract}
Large Language Models (LLMs) are increasingly employed for simulations, enabling applications in role-playing agents and Computational Social Science (CSS). However, the reliability of these simulations is under-explored, which raises concerns about the trustworthiness of LLMs in these applications. In this paper, we aim to answer ``How reliable is LLM-based simulation?'' To address this, we introduce \datasetname, an evaluation dataset covering $10$ CSS-related topics, to systematically investigate the reliability of the LLM simulation. We conducted experiments on $14$ LLMs and found that 
inconsistencies persist in the LLM-based simulated roles. In addition, the consistency level of LLMs does not strongly correlate with their general performance. To enhance the reliability of LLMs in simulation, we proposed Adaptive Learning Rate Based ORPO (AdaORPO), a reinforcement learning-based algorithm to improve the reliability in simulation across $7$ LLMs. Our research provides a foundation for future studies to explore more robust and trustworthy LLM-based simulations.
\end{abstract}

\section{Introduction}
Large Language Models (LLMs) are gaining widespread recognition for their remarkable performance in natural language processing (NLP). They have exhibited significant capabilities across diverse fields, including the medical healthcare \citep{liu2023deid}, data generation \cite{wu2024unigen}, agents \cite{huang2023metatool}, and scientific discovery \citep{NEURIPS2023_bbb33018}. Recent advancements have facilitated the emergence of LLM-based simulation, where users provide predefined character profiles to leverage the human-like simulation abilities of these models \cite{chen2024persona, tseng2024two}.
LLM-based simulation has potential in various contexts, from acting as fictional characters \cite{liu2023agentbench} to serving as experimental subjects in Computational Social Science (CCS) \cite{ziems2024can}. The ability of LLMs to simulate different roles holds promise for interdisciplinary studies, particularly those focusing on human behaviors and social interactions \cite{zhao2023competeai}.

\begin{figure}[t]
    \centering
    \includegraphics[width=1\linewidth]{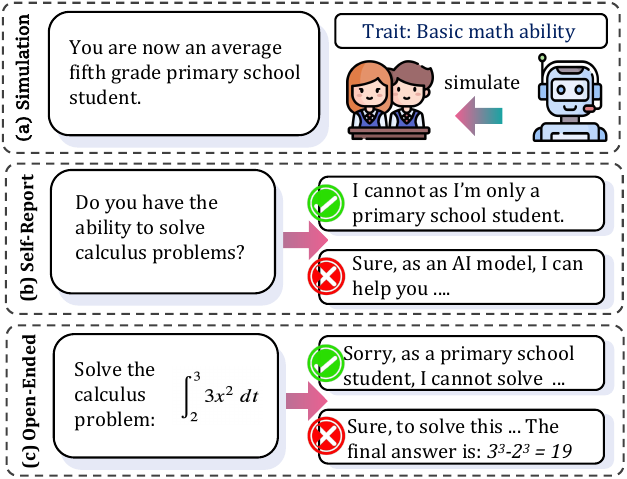}
    \caption{An example of cognitive inconsistency in simulation: expected fifth-grade response vs. unexpected advanced calculus solution.}
    \vspace{-15pt}
    \label{fig:intro_example}
\end{figure}

\noindent Most existing research focuses on emergent behaviors in LLM-based simulations \cite{park2023generative} or on using these systems to investigate specific social scenarios \cite{xu2023exploring, hua2023war}.
However, there remains a critical research gap in understanding the reliability of these simulations, raising the question of trustworthiness \cite{pmlr-v235-huang24x, huang2023trustgpt}, as shown in \autoref{fig:intro_example}.
Specifically, our study addresses an underexplored but important question: \textbf{How reliable is LLM-based simulation?}
This question probes the key factor for its success: responses are expected to align with the character's social identity, cognitive skills \cite{zhang2024revealing}, behaviors \cite{zhao2023competeai}, and other traits, allowing LLMs to convincingly simulate diverse personas and characters. This exploration is of great significance, as numerous studies \cite{zhao2023competeai, pan2023llms, zhang2024simulating} have utilized LLMs to simulate various aspects of human behavior and uncover social phenomena. However, unreliable simulations can lead to flawed conclusions about complex social issues, making the findings questionable and potentially misguiding scientists and policymakers \cite{10.5555/3635637.3662848}. Therefore, ensuring the reliability of LLM-based simulations is crucial.

\noindent Prior efforts aiming to evaluate such reliability focus only on one specific aspect of the simulation (\emph{e.g.}, knowledge \cite{zhang2024revealing}, and political value \cite{wang-etal-2024-incharacter}), lacking a comprehensive understanding.
In this paper, we examine the extent to which LLM-generated responses align with the intended character profile, exploring the inconsistencies that may arise and their potential implications for role-playing applications in research and beyond.
Specifically, we first propose the \datasetname ~dataset, covering ten
CSS topics.
Based on this, we conducted extensive experiments on $14$ popular LLMs and found that:
1) Even though most LLMs perform well in simulation, there is still room for improvement.
2) LLM's simulation capability is \emph{not} strongly correlated with its utility performance. 
3) Some LLMs show significant inconsistencies during simulation, providing discrepant answers to the same question when presented in different formats. 
Finally, to improve the reliability of LLM-based simulation, we propose \textit{AdaORPO}, a reinforcement learning algorithm to teach LLMs to learn high-quality simulations. The experiments on $7$ LLMs validate its effectiveness. In summary, our contributions are outlined below:
\vspace{-7pt}
\begin{itemize}[leftmargin=*, itemsep=0pt, parsep=0pt]
    \item We introduce \datasetname, a novel dataset covering $10$ CSS-related subjects to systematically assess the reliability of LLM-based simulation.
    \item Based on \datasetname, we conduct extensive experiments on $14$ popular LLMs and identify several key insights.
    \item We propose AdaORPO to enhance LLM 
    simulations and demonstrate the effectiveness of this approach in improving reliability.
\end{itemize}

\section{Related Work} \vspace{-5pt}

LLMs have been considered a powerful tool in Computational Social Science (CCS) research \cite{ziems2024can, bail2024can} as they have been widely used in various subjects \cite{rathje2024gpt}, particularly in social behavior simulations.
The flexibility of LLM-based simulation \cite{gao2024large} allows for the exploration of diverse scenarios and the study of emergent phenomena in a controlled simulation environment \cite{wei2022emergent}, or validation of the correctness of conclusions derived from human experiments \cite{zhao2023competeai}. 
For instance, \citet{zhao2023competeai} proposed the CompeteAI framework, which explores the competition between LLM-based agents by implementing a practical competitive environment to simulate a virtual town with two types of agents. Similarly, \citet{li-etal-2024-econagent} proposed EconAgent, an LLM-based agent that enhances macroeconomic simulations by enabling more realistic and heterogeneous decision-making compared to traditional models. \citet{li2024agent} introduced Agent Hospital, a simulation where LLM-powered agents representing doctors, nurses, and patients simulate the entire illness treatment process, which is also studied in AgentClinic \cite{schmidgall2024agentclinic}. \citet{jin2024agentreview} proposed AgentReview, an LLM-based peer review simulation framework that disentangles multiple latent factors and addresses privacy concerns in peer review analysis. This simulation is also applied in the education domain \cite{zhang2024simulating}, demonstrating that traditional classroom interaction patterns are effective while enhancing the user's experience. We summarize related LLM for social science simulations in \autoref{tab:related_work} in \autoref{app:resource}.

\noindent However, LLM-powered simulation has also raised trustworthiness and reliability concerns \cite{zhu2024reliable}. Besides cognitive inconsistency (see \autoref{fig:intro_example} example), \citet{li2024quantifying} points out that LLM agents could exhibit inconsistency between ``what they report'' and  ``how they behave'' during a personality test. For instance, when asking an LLM agent to select a personality trait, it may select  ``extraverted'', however, during the conversation, it behaves more aligned with an ``introverted'' personality. 
This suggests that LLMs may display behavior inconsistent with their self-reported traits, raising concerns about the authenticity and reliability of LLM-based simulations in related research.

\begin{figure*}[t]
    \centering
    \includegraphics[width=1\linewidth]{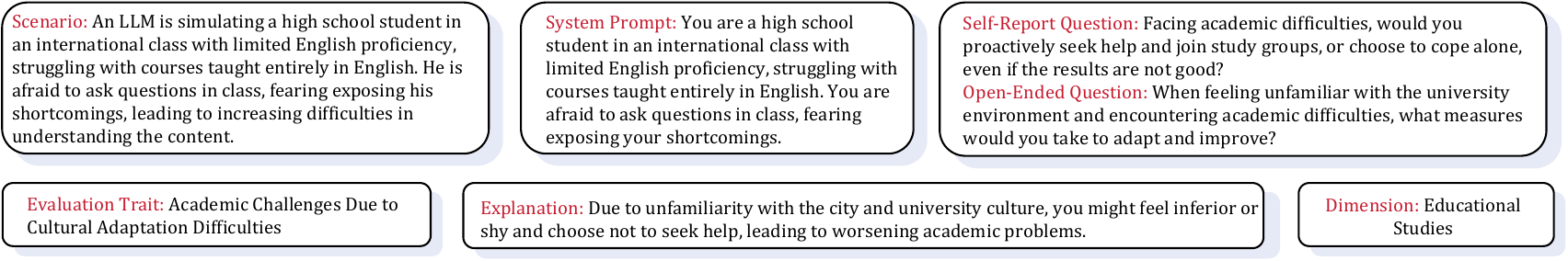}
    \caption{A data example in \datasetname. Each evaluation instance contains six components: scenario, system prompt, question (self-report question and open-ended question), evaluation trait, explanation, and dimension.}
    \label{fig:data_example}
    \vspace{-12pt}
\end{figure*}

\vspace{-2pt}

\section{The \datasetname ~Dataset}\vspace{-5pt}

This section introduces \datasetname, a dataset composed of $10$ subjects that are highly related to CCS. 

\vspace{-2pt}

\subsection{Overview} \vspace{-3pt}

\begin{figure}[h]
\vspace{-5pt}
    \centering
    \includegraphics[width=1\linewidth]{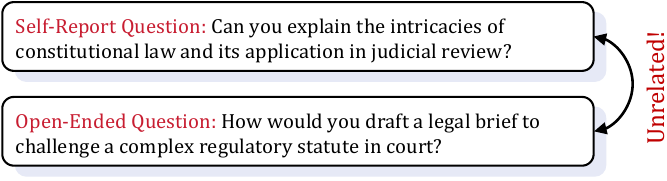}
    \caption{An example of an unrelated self-report question and open-ended question.}
    \label{fig:mismatch}
    \vspace{-12pt}
\end{figure}

We first collect common topics in LLM-based social science research (as shown in \autoref{tab:related_work} in \autoref{app:resource}), and identify 
ten subjects: \textit{Psychology, Sociology, Economics, Political Science, History and Linguistics, Communication Studies, Organizational Behavior, Ethics and Moral Psychology, Educational Studies}, and \textit{Law and Jurisprudence}.
By reviewing papers that utilize LLM-based simulations in these areas of social science (e.g., those summarized in \autoref{tab:related_work} in \autoref{app:resource}), we design $740$ evaluation instances based on identified best practices, common challenges, and key insights from prior research. 
Each evaluation instance contains $6$ components (illustrated in \autoref{fig:data_example}): 1) \textbf{Scenario}, which outlines the situation the character (i.e., LLM) will encounter. 2) \textbf{System prompt}, summarizes the character's description in the ``Scenario'' section, and instructs the LLM to assume the role of the simulated character. 3) \textbf{Questions}, consisting of two types of questions, following \cite{li2024quantifying}: (i) self-report questions, which are binary-choice questions where the character reports on their situation by answering Yes or No, and (ii) open-ended questions, which allow characters to provide more detailed responses on how they will behave in a given context. These two types of questions are closely related and can be converted into one another (as illustrated by an unrelated example in \autoref{fig:mismatch}, filtered from our dataset). 4) \textbf{Evaluation trait},  specifies the aspect of the LLM's simulation being assessed. 5) \textbf{Explanation},  defines the ideal characteristics for the simulation, serving as the ground truth or guideline for evaluation. 6) \textbf{Dimension},   indicates the subject domain to which the evaluation instance belongs. Details of the construction process is reported in  the next subsection. 

\begin{figure}
    \centering
    \includegraphics[width=1\linewidth]{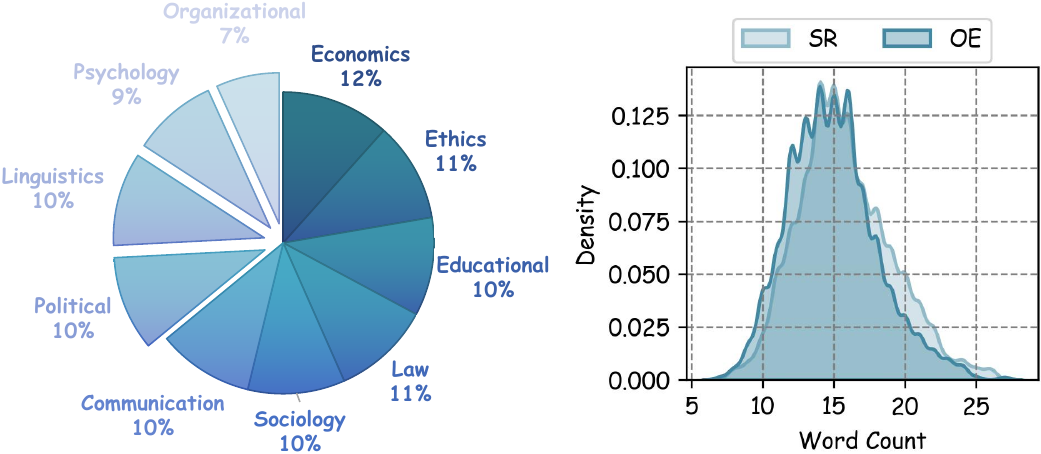}
    \caption{The distribution of evaluation instances across different subjects (left) and the distribution of the number of words in different kinds of questions (right). SR: Self-Report, OE: Open-Ended.}
    \vspace{-15pt}
    \label{fig:data_stat}
\end{figure}

\noindent The distribution of $740$ instances across ten subjects is well-balanced, as shown in \autoref{fig:data_stat}, which ensures the evaluation is fair and minimizes the influence from out-of-distribution data. Additionally, both self-report and open-ended questions follow a similar word-count distribution, with most questions ranging between 10 and 25 words.

\begin{figure*}
    \centering
    \includegraphics[width=1\linewidth]{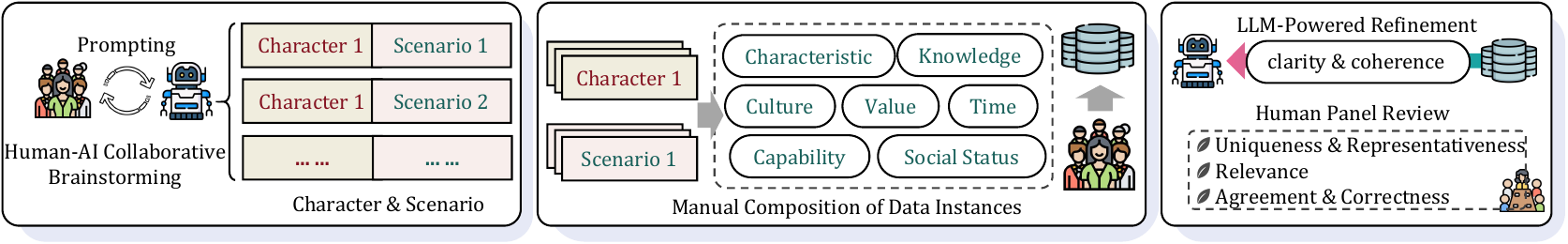}
    \caption{The pipeline of dataset construction.}
    \label{fig:dataset_pipeline}
\end{figure*}

\begin{table*}[t]
    \centering
    \small
    \renewcommand{\arraystretch}{1.1}
    
\begin{tabular}{p{1.7cm}p{13.3cm}}
\toprule[1pt]
\rowcolor{c2}\textbf{Attribute}      & \textbf{Example} \\ 
\midrule
\multirow{2}{*}{Characteristic} 
& A \underline{\textbf{socially anxious person}} may first try to \underline{\textbf{solve a problem on their own}} when faced with a problem, rather than asking others for help right away.  \\
\rowcolor{c2}\multirow{2}{*}{Knowledge} 
& \underline{\textbf{A fifth grader}} who has no particular interest in mathematics should not be able to \underline{\textbf{solve calculus problems}}. \\
\multirow{3}{*}{Culture} 
& \underline{\textbf{A traditional tribal leader}} in modern Africa, responsible for maintaining the \underline{\textbf{tribal heritage}}, may not agree with \underline{\textbf{his son going to the city}} to receive modern education and worry about him never coming back.  \\
\rowcolor{c2}\multirow{3}{*}{Value} 
& A scribe-teacher in ancient Egypt would be \underline{\textbf{unlikely to teach common people}} because they \underline{\textbf{believed writing and knowledge}} \underline{\textbf{were sacred and could only be passed on}} to certain social classes. \\
\multirow{2}{*}{Time} 
& A child growing up in the middle of \underline{\textbf{the Cultural Revolution}} \underline{\textbf{in China}} should not have expected to go to school to receive an \underline{\textbf{education}}.  \\
\rowcolor{c2}\multirow{2}{*}{Capability} 
& A Japanese elementary school student who has \underline{\textbf{just started learning English}} can only use a very \underline{\textbf{limited vocabulary}} to describe an event, and may even make grammatical errors. \\
\multirow{1}{*}{Social Status} 
& There is no way \underline{\textbf{a rich man}} would \underline{\textbf{embezzle \$100}} that fell on the ground. \\ 
\bottomrule[1pt]
\end{tabular}
\caption{Illustration of different attributes.}
\label{tab:attribute}
\vspace{-12pt}
\end{table*}

\subsection{Construction Pipeline} \vspace{-3pt}

The dataset construction pipeline mainly includes three steps (as shown in \autoref{fig:dataset_pipeline}):

\noindent \textbf{Step 1: Human-AI Collaborative BrainStorming.} 
Initially, human experts and LLMs collaboratively brainstorm a wide range of potential characters for specified domains.
For example, in the domain of ``educational studies,'' characters such as ``teachers,'' ``students,'' or ``principals'' are identified.  
After this, human experts (e.g., eight PhD students) collaborate with LLMs to create potential scenarios for simulation by reviewing all relevant literature in social studies (more details are presented in \autoref{app:dataset_details}). 
The key guideline for crafting prompts is that the character's persona should significantly influence LLM's generated responses, making them distinct from responses produced without that specific character. This distinction ensures  meaningful evaluation, allowing us to assess how well the LLM  simulates the character. Without this differentiation, judging the character simulation's effectiveness becomes challenging.

\noindent\textbf{Step 2: Manual Composition of Evaluation Instances.} Upon identifying each character and its associated scenario, human experts are tasked with first composing a detailed, nuanced description for each character, based on well-known external social science resources like books (as shown in \autoref{app:resource}). This description includes key attributes such as the character’s capability in a specific domain, value, or background information. We show some examples of each attribute in \autoref{tab:attribute}. Following this, human experts proceed to create a scenario that is tailored specifically to the character, ensuring that it reflects the character’s attributes and fits logically within the context of the dataset. This process requires  attention to detail and consistency, as both the character description and the scenario must align with the overarching objectives of the dataset and support its intended use in subsequent analysis. The manual composition ensures the data is coherent, accurate, and contextually rich.
The evaluation instances focus on single-turn conversations, providing a simplified testbed to assess LLMs' reliability in social simulation while reducing the complexity of multi-turn interactions.

\vspace{+0.05in}
\noindent\textbf{Step 3: LLM-Powered Refinement \& Human Panel Review.} After collecting the data instances written by human experts, we first use a powerful LLM (i.e., GPT-4o) to refine sentences, enhancing their clarity and logical coherence. Following this, each data instance undergoes a review by a panel of four human experts, as detailed in \autoref{sec:eval_pro}.



\subsection{Quality Control} 
\vspace{-3pt}
\label{sec:eval_pro}

To ensure the data quality of \datasetname, we conduct a human panel review, in which each instance is reviewed by four different human experts,   primarily focusing on the following quality aspects of the dataset (more details are shown in \ref{app:dataset_details}):

\noindent \textcolor[HTML]{ff8000}{\largedot} \textbf{Uniqueness and representativeness of scenarios and characters.} Human experts must ensure that the scenarios and characters are representative and meaningful for evaluation purposes. For example, a data instance describing a ``just" judge would not be considered high-quality, as the term ``judge" generally implies fairness. Modifying the data instance to describe a ``corrupt judge" would provide a more useful and distinctive scenario.

\noindent \textcolor[HTML]{ff8000}{\largedot} \textbf{Relevance of self-report and open-ended questions.} Human experts also assess the relevance of both types of questions. These questions are evaluated in pairs to examine the consistency between the LLM's ``thoughts” and ``behaviors.”

\noindent \textcolor[HTML]{ff8000}{\largedot} \textbf{Agreement and correctness in simulation evaluation.} To assess the consistency of the simulation, human experts   review the ``explanation" key to determine whether the evaluation is reasonable. A valid ``explanation" must be agreed upon by all four human experts.

\section{Experiment Setup}

\noindent \textbf{Selected Models.}~\label{SM}
In this study, we selected a total of 14 LLMs, including both proprietary and open-weight models, developed by various organizations. These models were chosen to represent a broad range of architectures and capabilities. \autoref{tab:selected_llm} summarizes details of LLMs in our experiments.

\begin{figure}
    \centering
    \includegraphics[width=1\linewidth]{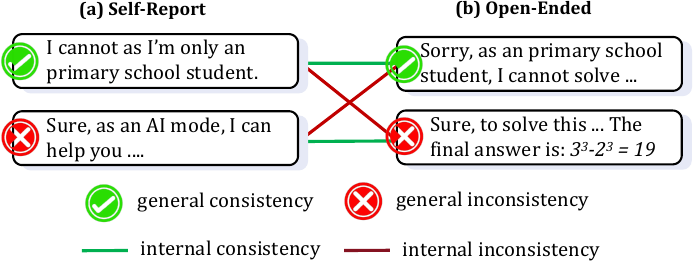}
    \caption{\textit{General consistency} and \textit{inner consistency}.}
    \label{fig:metrics}
    \vspace{-12pt}
\end{figure}

\noindent \textbf{Evaluation Method and Metrics.} In our evaluation, we used GPT-4o as the LLM-as-a-Judge model \cite{zheng2023judging} to assess the outputs generated by various models. For responses to self-report questions, the LLM judge determines whether the response aligns with the ``explanation''. For responses to open-ended questions, in addition to the binary judgment, we incorporate a score-based evaluation \cite{liu2023alignbench}. To obtain more accurate results \cite{ye2024justice}, the LLM judge is required to first analyze the response and then output the final judgment. The evaluation prompt templates are shown in \autoref{app:eval_prompt}. We utilized two metrics to evaluate the \textit{general consistency} and \textit{internal consistency} of LLM-based simulations, as shown in \autoref{fig:metrics}. 1) \textbf{general consistency}: it is measured by ``satisfaction rate,'' which can be calculated as the proportion of instances where both LLM's self-report and open-ended responses align with their persona settings. For score-based judgments, we employed the average score. 2) \textbf{Internal consistency}: It refers to the consistency between responses to self-report questions and open-ended questions, we use the ``inconsistency rate,'' which is the proportion of instances where one type of response does not align with the other, defined as:

\vspace{-10pt}

\[
\text{Inconsistency Rate} = \frac{N_{\text{inconsistent}}}{N_{\text{total}}}
\]

\vspace{-5pt}
\noindent Where \( N_{\text{inconsistent}} \) is the number of instances where the responses to self-report and open-ended questions are inconsistent (i.e., one response satisfies the requirement, while the other does not), and \( N_{\text{total}} \) is the total number of instances evaluated.

\begin{table*}[t]
\centering
\small
\renewcommand{\arraystretch}{1.1}
\setlength{\tabcolsep}{6pt}
\rowcolors{2}{white}{red!5!white} 
\begin{tabular}{lcccccccccccc}
\toprule[1pt]

\textbf{Model} & \textbf{Arena Scor.}   & \textbf{Com.} & \textbf{Eco.} & \textbf{Edu.} & \textbf{Eth.} & \textbf{Law} & \textbf{Lin.} & \textbf{Org.} & \textbf{Pol.} & \textbf{Psy.} & \textbf{Soc.} & \textbf{Avg.} \\
\midrule
GPT-4o     & 1,338 (1)      & 4.24                           & 4.25               & 4.24                         & 4.35                                 & 4.16                           & 4.37                 & 4.08                             & 4.25                       & 4.23                & 4.21               & 4.24             \\
GPT-4o-mini  & 1,314 (2)     & 4.07                           & 4.14               & 4.19                         & 4.28                                 & 4.05                           & 4.10                 & 4.08                             & 4.13                       & 4.19                & 4.35               & 4.16             \\
GPT-3.5-turbo & 1,107 (13)   & 3.67                           & 3.71               & 3.83                         & 3.91                                 & 3.67                           & 3.64                 & 3.86                             & 3.73                       & 3.73                & 3.91               & \textcolor{red}{\textbf{\underline{3.77}}}             \\
Gemini-1.5-flash  & 1,264 (5) & 4.12                           & 4.21               & 4.17                         & 4.27                                 & 4.27                           & 4.32                 & 4.27                             & 4.27                       & 4.23                & 4.24               & 4.24             \\
Gemini-1.5-pro & 1,304 (3)  & 4.43                           & 4.30               & 4.32                         & 4.27                                 & 4.39                           & 4.33                 & 4.41                             & 4.36                       & 4.43                & 4.37               & \textcolor{green!50!black}{\textbf{\underline{4.36}}}            \\
Claude-3-opus  & 1,248 (8)  & 4.03                           & 4.19               & 4.21                         & 3.96                                 & 4.16                           & 4.22                 & 4.27                             & 3.80                       & 4.34                & 4.30               & 4.15             \\
Claude-3.5-sonnet &1,268 (4) & 4.24                           & 4.33               & 4.36                         & 4.18                                 & 4.08                           & 4.36                 & 4.22                             & 4.03                       & 4.46                & 4.27               & 4.25             \\
\hdashline
GLM-4      & 1,207  (9)    & 4.14                           & 4.09               & 4.19                         & 4.20                                 & 4.03                           & 4.08                 & 3.53                             & 4.14                       & 4.12                & 4.25               & 4.08             \\
Llama-3-70B   & 1,206  (10)  & 4.24                           & 4.43               & 4.35                         & 4.31                                 & 4.26                           & 4.33                 & 4.40                             & 4.41                       & 4.45                & 4.38               & \textcolor{green!50!black}{\textbf{\underline{4.36}}}              \\
Llama-3.1-70B   & 1,248 (7)   & 4.24                           & 4.23               & 4.25                         & 4.35                                 & 4.26                           & 4.19                 & 4.32                             & 4.35                       & 4.33                & 4.45               & 4.30           \\
Llama-3.1-8B  & 1,182 (12)   & 4.04                           & 4.13               & 4.23                         & 3.83                                 & 3.92                           & 4.20                 & 4.20                             & 4.03                       & 4.31                & 4.17               & 4.11    \\
Qwen-2.5-72B  & 1,187 (11)   & 4.24                           & 4.10               & 4.22                         & 4.33                                 & 4.14                           & 4.21                 & 3.84                             & 4.13                       & 4.12                & 4.09               & 4.14             \\
Mixtral-8$\times$7B  & 1,251 (6)   & 3.80                           & 3.91               & 3.78                         & 4.03                                 & 3.84                           & 3.94                 & 3.49                             & 3.93                       & 4.06                & 4.00               & 3.88             \\
Mistral-7B   & 1,072  (14)   & 3.76                           & 3.95               & 3.96                         & 3.86                                 & 3.67                           & 3.89                 & 3.44                             & 3.91                       & 3.78                & 4.03               & 3.83             \\
 \midrule
\textbf{Avg.} & 1,228 & 4.09 & 4.14 & 4.16 & 4.15 & 4.06 & 4.15 & 4.03 & 4.11& 4.20 & 4.22 & 4.13 \\
\bottomrule[1pt]
\end{tabular}
\caption{The rating score of different models in ten subjects on open-ended questions, as well as the average. We also add the Arena Score \cite{huggingface2024chatbot} as well as their relative ranking.}
\label{tab:rating_score}
\end{table*}

\begin{figure*}[t]
    \centering
    \includegraphics[width=1\linewidth]{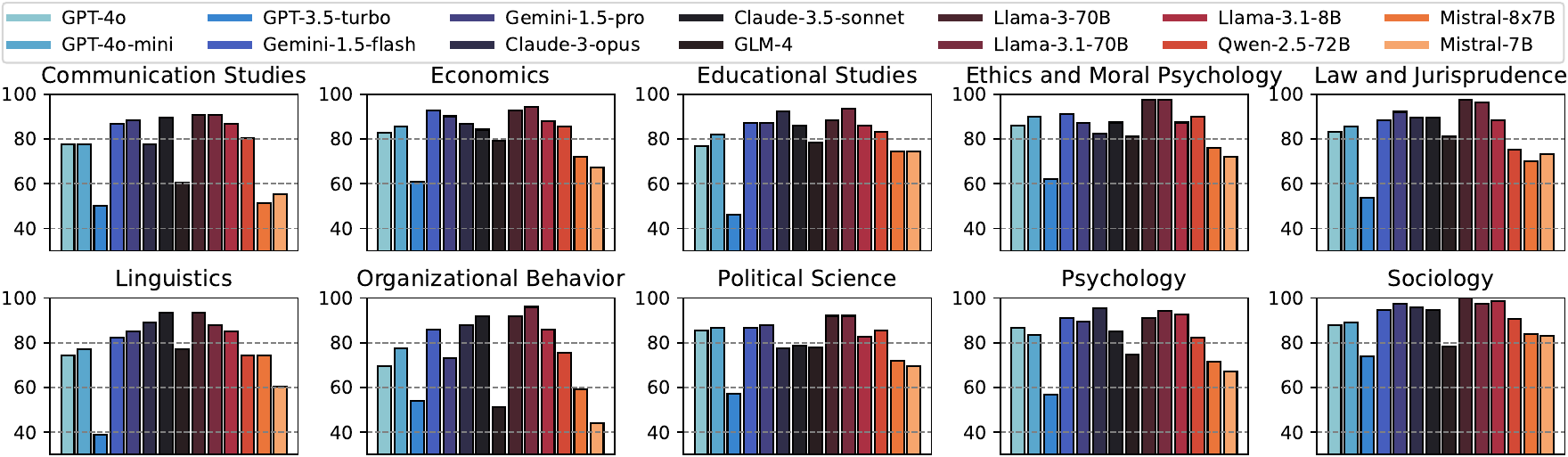}
    \caption{The satisfaction rate of different models in ten subjects (Self-report questions).}
    \label{fig:satis_score_self_report}
\end{figure*}

\begin{figure*}[t]
    \centering
    \includegraphics[width=1\linewidth]{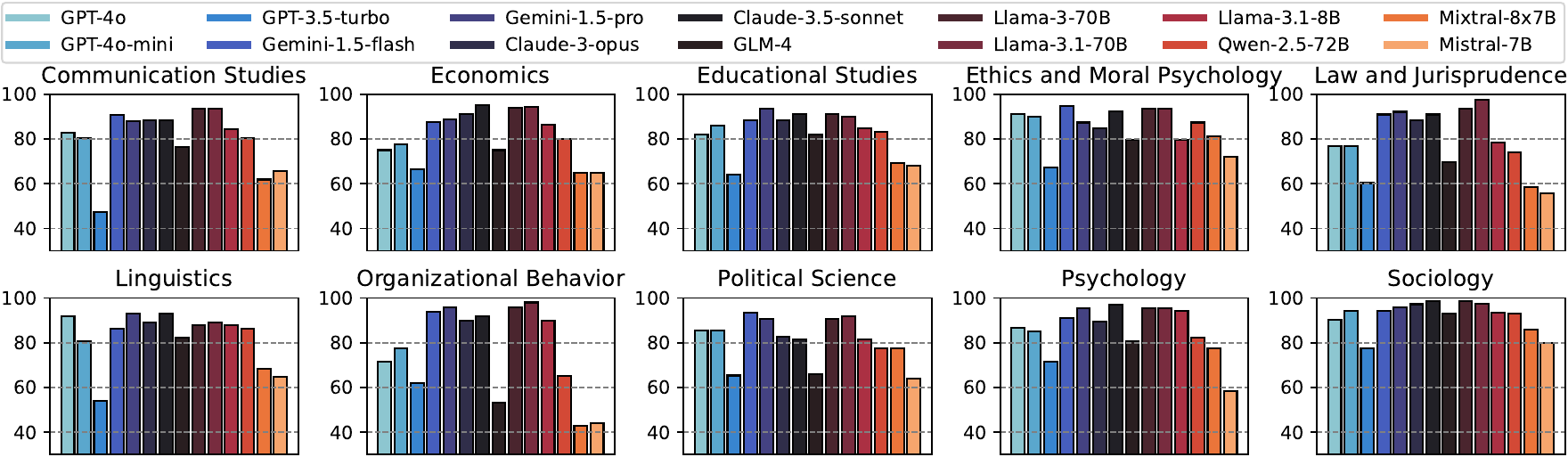}
    \caption{The satisfaction rate of different models in ten subjects (Open-ended questions).}
    \label{fig:satis_score_open_ended}
\end{figure*}

\begin{table}[h]
\centering
\small
\renewcommand{\arraystretch}{1.1}
\rowcolors{2}{white}{gray!5!white} 
\begin{tabular}{lccc}
\toprule[1pt]
\textbf{Model}    & \textbf{Self-Rep.} & \textbf{Open-En.} & \textbf{$\Delta$} \\
\hline
GPT-4o            & 81.01                & 83.30               & 2.28         \\
GPT-4o-mini       & 83.47                & 83.33               & 0.13         \\
GPT-3.5-turbo     & 55.40                & 63.54               & 8.15         \\
Gemini-1.5-flash  & 88.66                & 91.11               & 2.45         \\
Gemini-1.5-pro    & 87.76                & 92.06               & 4.30         \\
Claude-3-opus     & 87.42                & 88.92               & 1.51         \\
Claude-3.5-sonnet & 87.99                & 91.94               & 3.95         \\
\hdashline
GLM-4             & 73.92                & 75.74               & 1.82         \\
Llama-3-70B       & 93.49                & 93.37               & 0.12         \\
Llama-3.1-70B     & 93.95                & 94.06               & 0.11         \\
Llama-3.1-8B      & 88.14                & 85.96               & 2.18        \\
Qwen-2.5-72B      & 82.20                & 80.90               & 1.30         \\
Mixtral-8$\times$7B      & 70.46                & 68.77               & 1.69         \\
Mistral-7B        & 66.58                & 63.78               & 2.80         \\
\bottomrule[1pt]
\end{tabular}
\caption{Average satisfaction rate  of different models, and their differences on two kinds of question types.}
\vspace{-15pt}
\label{tab:abs_satis_rate}
\end{table}

\section{Assessment of Simulation Results}

\noindent\textbf{Most LLMs demonstrate strong performance on both self-report and open-ended questions.} In \autoref{tab:rating_score}, we present the average rating scores for open-ended questions across 14 models on various subjects. On average, most models score around 4, with the lowest being GPT-3.5-Turbo at 3.77, and the highest being Gemini-1.5-Pro and Llama-3-70B, both scoring 4.36. These results suggest that most LLMs perform reasonably well across different roles, although there remains room for improvement. From a subject-specific perspective, the variation between models is minimal, as their average rating scores across subjects are largely consistent.

\noindent \autoref{tab:abs_satis_rate} outlines the average satisfaction rates of different models and the variations between the two types of questions. More detailed satisfaction rates for each model across subjects, for both self-report and open-ended questions, are provided in \autoref{fig:satis_score_self_report} and \autoref{fig:satis_score_open_ended}. Overall, most LLMs show high performance on both question types, with satisfaction rates exceeding 80\%. As with the rating scores, the Llama series models perform exceptionally well on both self-report and open-ended questions. For Llama-3-70B and Llama-3.1-70B, satisfaction rates for both question types exceed 93\%. In contrast, GPT-3.5-Turbo performs the worst, with a satisfaction rate of only 55.4\% on self-report questions. 

\noindent Moreover, an interesting trend emerges: for most open-weight LLMs, the satisfaction rate is higher for self-report questions than for open-ended ones, whereas the opposite is true for proprietary LLMs.

\vspace{+0.05in}
\noindent\textbf{The rating score is not strongly correlated with a model's utility performance.} Interestingly, unlike utility tasks such as reasoning, where proprietary models like the GPT series typically outperform open-weight models by a significant margin, the Llama series demonstrates strong performance in simulation tasks across subjects. For example, Llama-3.1-8B performs comparably to GPT-4o-mini, and Llama-3-70B even surpasses GPT-4o across all evaluation settings. Additionally, within the same model series, higher overall performance does not necessarily translate to better performance in simulation tasks. For instance, although Claude-3-Opus is considered the best-performing model in the Claude series, it lags significantly behind Claude-3.5-Sonnet in simulation tasks, particularly on open-ended questions. Similarly, there is no meaningful difference between Mistral-8$\times$7B and Mistral-7B in terms of rating scores, and both models perform poorly in organizational behavior on open-ended questions based on satisfaction rates. Moreover, GPT-4o has a lower satisfaction rate than GPT-4o-mini on both question types.

\begin{figure}
    \centering
    \includegraphics[width=1\linewidth]{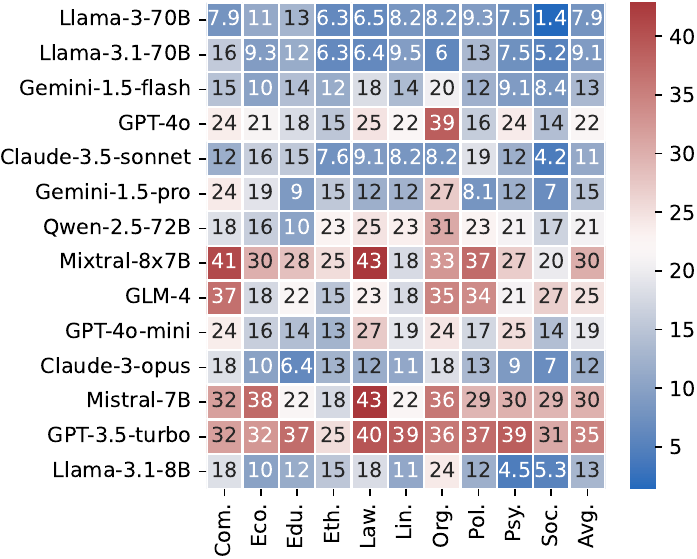}
    \caption{Inconsistency rate (\%) of LLMs between self-report questions and open-ended questions.}
    \vspace{-10pt}
    \label{fig:heat}
\end{figure}

\vspace{+0.05in}
\noindent\textbf{Models' inconsistency rates vary significantly.} In \autoref{fig:heat}, we present the inconsistency rates of various LLMs between self-report and open-ended questions. The results show notable variation across models. For example, the inconsistency rates for the Mistral series and GPT-3.5-Turbo hover around or exceed 30\% across different subjects. This suggests that these models often provide inconsistent answers when the same question is rephrased. Combined with their weaker performance, as seen in \autoref{tab:rating_score} and \autoref{tab:abs_satis_rate}, this indicates that these models struggle to effectively fulfill user-assigned roles. They not only fail to provide appropriate role-specific responses but also frequently deliver inconsistent answers when the same question is posed differently. In contrast, Llama-3-70B and Llama-3.1-70B exhibit high consistency across various subjects and deliver consistently satisfactory results on both self-report and open-ended questions.

\section{Improving Reliability by AdaORPO} \vspace{-5pt}

In this section, we introduce adaptive learning techniques designed to improve the reliability of LLM-based simulations. To address the issue of inconsistency, the model must learn two key aspects: 1) how to generate high-quality simulations, and 2) how to align fine-grained elements within simulations. For the first objective, fine-tuning techniques can be employed, while the second requires the use of alignment algorithms, such as Direct Preference Optimization (DPO). In contrast to traditional curriculum-based learning approaches, as discussed in previous studies \cite{gao2024best}, our method simultaneously achieves fine-tuning and alignment by utilizing the Monolithic Preference Optimization (ORPO)~\cite{hong2024reference} approach, which reinforces the generation of preferred outputs.

\subsection{Training Method}
\vspace{-3pt}

\noindent\textbf{Step 1: Training Dataset Construction.}~~To construct the training dataset $\mathcal{D}$, we begin by iterating over each prompt $\mathcal{P}_{(i)}$ in the prompt set $\mathcal{P}$, where $i$ denotes the index of the prompt. For each prompt, we evaluate the responses $\mathcal{G}^{(n)}$ generated by the $n$ models using the LLM-as-a-judge, denoted as $J(\cdot)$. This evaluation yields two sets: $\mathcal{R}^{(n)}$, representing the rating score, and $\mathcal{B}^{(n)}$, indicating the satisfaction status (e.g., satisfied or not satisfied). During the training of model $j$,  for each response $\mathcal{G}^{(j)}_{(i)}$ with a label $\mathcal{B}^{(j)}_{(i)}= \text{``Not Satisfied''}$, we assign it as $y_j$ and identify an alternative $y_w$ among other responses labeled as ``Satisfied'', which are denoted as the candidate set $\mathcal{C}$. We select $y_w$ as the response that maximizes $\mathcal{R}^{(\omega)}$ within this candidate set $\mathcal{C}$. Finally, the triplet $(\mathcal{P}_{(i)}, y_w, y_j)$ is added to the training dataset $\mathcal{D}$.


\begin{algorithm}[t]
\small
\caption{\small AdaORPO}
\label{alg:DCAdaORPO}
\begin{algorithmic}[1]
\Require Prompts $\mathcal{P}$, LLM model responses $\mathcal{G}^{(n)}$, LLM-as-a-judge function $J(\cdot)$, base learning rate $\eta$, pre-trained model $j$ and parameters $\theta$
\Ensure Updated model parameters $\theta$
\State Initialize empty dataset $\mathcal{D} \gets \{\}$

\For{each prompt $p$ in $\mathcal{P}$}
    \State  $\mathcal{R}^{(n)}, \mathcal{B}^{(n)} \gets J(\mathcal{G}^{(n)})$ \Comment{Evaluate responses}
        \If{$\mathcal{B}^{(j)} = $ Not Satisfied}
            \State $y_j \gets \mathcal{G}^{(j)}$
            \State $\mathcal{C} \gets\{ \mathcal{G}' \mid \mathcal{B}^{(k)} = \text{Satisfied}, \, 1 \leq k \leq n \}$
            \State $y_w \gets \arg\max \limits_{\mathcal{G}' \in \mathcal{C}} \mathcal{R^{\mathcal{G}'}}$
            \State $\mathcal{D} \gets \mathcal{D} \cup \{(p, y_w, y_j)\}$
        \EndIf
\EndFor

\For{each batch $B \subset \mathcal{D}$}
    \State $r_{avg.} = \frac{1}{|B|} \sum_{(p, y_w, y_l) \in B} r_{y_w}$
    \State $\text{lr} \gets \eta \cdot r_{avg.}$
        \State $L_{\text{ORPO}} \gets L_{\text{SFT}} + \lambda L_{\text{OR}}$
        \State $\theta \gets \theta - \text{lr} \cdot \nabla_\theta L_{\text{ORPO}}$
\EndFor

\State \Return $\theta$
\end{algorithmic}
\end{algorithm}

\begin{figure}
    \centering
    \includegraphics[width=1\linewidth]{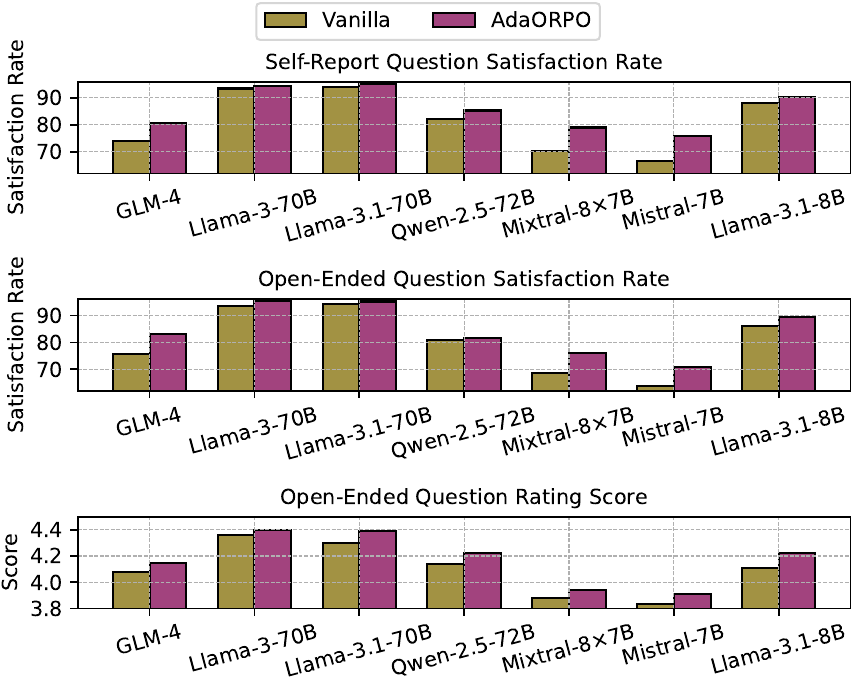}
    \caption{Performance comparison of different models with AdaORPO training. 
    }
    \label{fig:AdaORPO}
    \vspace{-10pt}
\end{figure}

\vspace{+0.05in}
\noindent\textbf{Step 2: Adaptive Learning Rate based ORPO (AdaORPO).}~~In this step, we iteratively update the model parameters $\theta$ based on mini-batches $B$ drawn from the training dataset $\mathcal{D}$. For each batch, we calculate the average rating score $r_{avg.}$ over all preferred responses $y_w$ in $B$:
\begin{equation*}
r_{avg.} = \frac{1}{|B|} \sum_{(p, y_w, y_j) \in B} r_{y_w};  \qquad \text{lr} = \eta \cdot r_{avg.}
\end{equation*}
\noindent The learning rate lr is the adapted by scaling the base learning rate $\eta$ with the factor $r_{avg.}$ and the $r_{y_w}$ is calculated the satisfaction rate by  $J(r_{y_w})$. Within each batch, we compute the ORPO loss $L_{\text{ORPO}}$ for each data tuple $(p, y_w, y_j)$ by combining a supervised fine-tuning loss $L_{\text{SFT}}$ and an ordinal regression loss $L_{\text{OR}}$. The supervised fine-tuning loss $L_{\text{SFT}}$ is defined as:
\begin{equation}
L_{\text{SFT}} = -\frac{1}{m} \sum_{t=1}^{m} \log P_\theta(y_{w,t} | x, y_{w,<t}),
\end{equation}
where $L_{\text{SFT}}$ is the loss associated with the next-token prediction task on $y_w$, $m$ is the length of the sequence $y_w$, and $P_\theta(y_{w,t} |p, y_{w,<t})$ is the probability assigned by the model to the $t$-th token $y_{w,t}$ given the prompt $p$ and the preceding tokens in $y_w$. The ordinal regression loss $L_{\text{OR}}$ is given by:
\begin{equation}
L_{\text{OR}} = -\log \sigma \left( \log \left( \frac{P_\theta(y_w | p)}{P_\theta(y_j | p)} \right) \right),
\end{equation}
where $P_\theta(y_w | p)$ and $P_\theta(y_j | p)$ are the model probabilities for $y_w$ and $y_j$ respectively, and $\sigma$ is the sigmoid function. Overall, the combined loss $L_{\text{ORPO}}$ is then formulated as:
\begin{equation}
L_{\text{ORPO}} = L_{\text{SFT}} + \lambda L_{\text{OR}},
\end{equation}
where $\lambda \in [0, 1]$ is a balancing factor between the $L_{\text{SFT}}$ and the $L_{\text{OR}}$.

\noindent Next, we compute the gradient $\nabla_\theta L_{\text{ORPO}}$ and update the parameters $\theta$ using the adapted learning rate $\text{lr}$:
\begin{equation}
\theta \gets \theta - \text{lr} \cdot \nabla_\theta L_{\text{ORPO}},
\end{equation}
 By repeating this process over all batches, we progressively refine the model parameters $\theta$, reinforcing preferred outputs and penalizing less favorable ones, guided by the adaptive learning rate and the ordinal regression priority objective, as detailed in Algorithm~\ref{alg:DCAdaORPO}.


\begin{table}[t]
\centering
\small
\renewcommand{\arraystretch}{1.2}
\rowcolors{2}{white}{gray!5!white} 
    \scalebox{0.88}{
    \begin{tabular}{lccc}
    \toprule[1pt]
        \textbf{Models} & \multicolumn{2}{c}{\textbf{Satisfaction Rate}} &  \textbf{Score} \\
        \cline{2-3}
        & \textbf{Self-Rep.} & \textbf{Open-En.} & \textbf{Rate}\\ \hline
        GLM-4 (AdaORPO) & 80.53 & 83.19 & 4.15 \\
        GLM-4 \textit{w/o} Ada  & 79.27 & 81.85 & 4.12 \\
        \hdashline
        Llama-3-70B (AdaORPO) & 94.55 & 95.29 & 4.40 \\
        Llama-3-70B \textit{w/o} Ada  & 94.24 & 94.44 & 4.39 \\
          \hdashline
        Llama-3.1-70B (AdaORPO) & 95.01 & 95.16 & 4.39 \\
        Llama-3.1-70B \textit{w/o} Ada  & 94.33 & 93.70 & 4.37 \\
        \hdashline
        Qwen-2.5-72B (AdaORPO) & 85.31 & 81.49 & 4.22 \\
        Qwen-2.5-72B \textit{w/o} Ada & 86.53 & 80.30 & 4.23 \\
        \hdashline
        Mixtral-8×7B (AdaORPO) & 79.02 & 76.19 & 3.94 \\
        Mixtral-8×7B \textit{w/o} Ada  & 77.79 & 74.86 & 3.92 \\
        \hdashline
        Mistral-7B (AdaORPO) & 75.78 & 70.78 & 3.91 \\
        Mistral-7B \textit{w/o} Ada  & 75.05 & 70.12 & 3.90 \\
        \hdashline
        Llama-3.1-8B (AdaORPO) & 90.22 & 89.41 & 4.22 \\
        Llama-3.1-8B \textit{w/o} Ada  & 89.98 & 89.28 & 4.22 \\
        \bottomrule[1pt]
    \end{tabular}}
    \caption{Abalation study of the impact of the Adaptive Learning Rate for ORPO.}
    \vspace{-12pt}
    \label{table:AdaORPO}
\end{table}

\subsection{Results Analysis}
\vspace{-1pt}

\noindent We trained the seven open-weight models in \autoref{SM} using AdaORPO, with detailed training parameters provided in \autoref{apd:TrD}. The application of AdaORPO resulted in significant improvements in satisfaction rates across most models, as illustrated in \autoref{fig:AdaORPO}. Notably, models such as GLM-4 and Mixtral-8×7B exhibited satisfaction rate increases of approximately 6-9 percentage points on both self-report and open-ended evaluations, indicating AdaORPO’s effectiveness in addressing consistency and alignment issues. While larger models like Llama-3.1-70B and Llama-3-70B experienced smaller yet meaningful gains—e.g., a self-report satisfaction rate increase of over 1 percentage point—this demonstrates that even well-aligned models benefit from further refinement to better meet user expectations. Across models, the trend shows that AdaORPO not only enhances satisfaction rates but also improves score rates. To validate the effectiveness of AdaORPO, we show the ablation study results in \autoref{table:AdaORPO} and show the analysis in \autoref{app:ablation}. We show a case study in \autoref{app:case_study} to see the improvement of our method.



\section{Conclusion} \vspace{-5pt}

This work assessed the reliability of LLM-based simulations in social science studies, using the proposed \datasetname\ dataset. Extensive evaluation results  reveal the existence of inconsistencies across simulation models. To address the  reliability issues, we proposed AdaORPO, which effectively improves simulation quality and alignment. Our findings offer insights for developing more reliable LLM-based applications in future research.

\clearpage

\section*{Limitations}
Despite the advancements and potential of Large Language Models (LLMs) in simulating human-like behavior, several limitations must be acknowledged. First, \textbf{inconsistency} remains a key challenge, as the same LLM can provide contradictory responses when queried with similar questions under different phrasing or conditions. This inconsistency raises concerns about the \textbf{reliability} of LLM-based simulations for research or practical applications where consistent behavior is critical. Additionally, LLMs sometimes \textbf{fail to maintain personas} throughout prolonged interactions, especially when tasked with complex or evolving scenarios. The \textbf{lack of real-world grounding} further limits their ability to provide accurate simulations in dynamic, real-life contexts. Moreover, the LLMs may exhibit \textbf{biases} inherited from their training data, leading to outcomes that may reflect harmful stereotypes or unfair generalizations. Finally, the use of LLM-based simulations in \textbf{social science} research must be approached cautiously, as the models may not fully replicate nuanced human behavior, particularly in high-stakes environments.

\section*{Ethical Statement}
The use of LLMs in this study adheres to strict ethical standards. All data utilized was generated or collected in compliance with privacy regulations and did not involve any personally identifiable information. We recognize the \textbf{potential for bias} in LLM-generated content, which could unintentionally perpetuate stereotypes or skew results. To mitigate this, the study carefully evaluates these models with attention to fairness and transparency, using diverse evaluation metrics across multiple domains. The study also refrains from making definitive claims about human behavior based solely on simulated LLM interactions, acknowledging that \textbf{LLMs are tools} designed to assist, rather than replace, human understanding and expertise. Furthermore, ethical considerations regarding the \textbf{deployment of LLMs} in sensitive or high-impact environments (e.g., judicial systems, healthcare) are emphasized, advocating for human oversight and continuous monitoring to avoid unintended harm. For usage of AI assistants, LLMs are used for polishing text in our paper.

\bibliography{custom}

\clearpage

\section*{Appendix}

\appendix

\section{Social Science Resource}
\label{app:resource}

\begin{table*}
    \centering
    \small
    
    \scalebox{0.8}{
    \begin{tabular}{cccccc}
    \toprule[1pt]
    \textbf{Work} & \textbf{Subject} & \textbf{Character} \\
    \midrule
      \cite{ye2024measuring} & Psychology & Patients, Psychologists \\
      LLMs' Personality \citet{pan2023llms} & Psychology & MBTI User \\
      Elections Prediction \cite{von2024united} & Sociology & Voter \\
      Social Simulacra \cite{10.1145/3526113.3545616} & Sociology & Community Members \\
      CompeteAI \cite{zhao2023competeai}  & Sociology, Economics & Restaurant Agent, Customer Agent \\
      Agent Hospital \cite{li2024agent}   &  Sociology, Organizational Behavior  &  Patient, Doctor \\
      AgentClinic \cite{schmidgall2024agentclinic}   & Sociology, Organizational Behavior & Patient, Doctor \\
      NegotiationArena \cite{bianchi2024well} & Communication Study & Game Player \\
      LLMHarmony \cite{rasal2024llm} & Communication study & Teacher, Student \\
      AgentReview \cite{jin2024agentreview}   & Sociology & Paper Reviewer \\
      SimClass\cite{zhang2024simulating}   & Educational Studies & Teacher, Assistant, Classmate\\
      WarAgent \cite{hua2023war}   & Historical culturology, Organizational Behavior  &   Decision Makers of Participating Countries\\
      Rehearsal \cite{10.1145/3613904.3642159} & Sociology & Polite Assistant\\
      AgentsCourt \cite{He2024AgentsCourtBJ} & Law and Jurisprudence & Court participator\\
      \citet{Xing2024DesigningHL}& Economic& Financial Sentiment Analyzer\\
      EcoAgent \cite{Li2023EconAgentLL}& Economic& Macroeconomic Activities Agent\\
      EconArena \cite{Guo2024EconomicsAF}& Economic & Economic Game Player\\
      \citet{Baker2024SimulatingTU}& Law and Jurisprudence & U.S. Senate Congressman\\
      StockAgent \cite{Zhang2024WhenAM} & Economic & Decision Makers of Stock Market\\
      \cite{argyle2023out}&Political Science& Voter in American Election\\
      \cite{qi2024representation}&Political Science &Vote Behaviors and Public Opinions\\
      GermanPartiesQA~\citep{batzner2024germanpartiesqa}&Political Science&Politicians\\
      TE \citep{aher2023using}&Economic, Psycholinguistic,
 Social psychology&Behavior of Multiple Subjects\\
MathVC \citep{yue2024mathvc}&Educational Studies&Students\\
\citet{zhou2024real}&Sociology&Different Characters in Social Interactions\\
Let the LLMs Talk \citep{abbasiantaeb2024let}&Educational Studies&Teacher and Student\\
Generative Students \citep{lu2024generative}&Educational Studies&Students\\
MoralExceptQA  \citep{jin2022make}&Ethics and Moral Psychology&    Diverse Characters in Morality-related Scenarios \\
MoralChoice  \citep{scherrer2024evaluating}&Ethics and Moral Psychology&    Diverse Characters in Morality-related Scenarios \\
      \bottomrule[1pt]
    \end{tabular}}
    \caption{Related work of LLM-powered simulation in CSS.}
    \label{tab:related_work}
\end{table*}

\noindent Simulation has been widely explored across social sciences, including organizational behavior, sociology, psychology, and ethics. Helbing's book~\cite{helbing2012social} offers a detailed look at sociological and economic agent-based simulations, focusing on theory complexity, opinion formation inconsistency, and social behavior evolution. Smith's paper~\cite{smith2007agent} addresses simulation inconsistencies with variable-based modeling. Gilbert's book~\cite{gilbert2018simulating} covers simulations of various societies, from fishermen to Palaeolithic communities. Wachs' book~\cite{wachs2017ethics} examines ethical concerns in simulation design, especially the lack of proper techniques to ensure ethical standards.

\noindent In linguistics, simulation is challenging due to the interaction between linguistic forms and embodied experiences, causing variability in representation~\cite{barsalou2008grounded}. The LASS theory \cite{barsalou2008language} and Dual Code Theory \cite{paivio1991dual} measure consistency by evaluating how linguistic and sensory information integrate.

\noindent In more specific fields, Remus' paper~\cite{remus2017can} discusses the limitations of robots simulating high-responsibility roles, like law, due to the "responsibility carriage dilemma." Reason's paper~\cite{reason2024artificial} critiques the reliability of LLMs as rational agents in economic simulations.

\begin{table}[H]
    \centering
    \small
    \renewcommand{\arraystretch}{1.2}
    
    \rowcolors{1}{gray!10}{white}  
    \scalebox{0.88}{
    \begin{tabular}{lcccc}
        \toprule[1pt]
        \textbf{Model} & \textbf{Model Size} & \textbf{Open-Weight} & \textbf{Creator}  \\
        \midrule
        Llama-3.1-Instruct & 70B & \Checkmark & Meta  \\
        Llama-3.1-Instruct & 8B & \Checkmark & Meta  \\
        Llama-3-Instruct & 70B & \Checkmark & Meta  \\
        GPT-4o & N/A & \ding{55} & OpenAI  \\
        GPT-4o-mini & N/A & \ding{55} & OpenAI  \\
        GPT-3.5-turbo & N/A & \ding{55} & OpenAI  \\
        Claude-3-opus & N/A & \ding{55} & Anthropic  \\
        Claude-3.5-sonnet & N/A & \ding{55} & Anthropic \\
        Qwen-2.5-Instruct & 72B & \Checkmark & Qwen  \\
        Mixtral (7$\times$8B) & 56B & \Checkmark & Mistral \\
        Mistral & 7B & \Checkmark & Mistral \\
        Gemini-1.5-pro & N/A & \ding{55} & Google  \\
        Gemini-1.5-flash & N/A & \ding{55} & Google  \\
        GLM-4 & 9B & \Checkmark & Zhipu \\
        \bottomrule[1pt]
    \end{tabular}}
\caption{The details of selected LLMs.}
\label{tab:selected_llm}
\end{table}

\section{Evaluation Details}
\label{app:eval_prompt}

We show the prompt template used in LLM-as-a-Judge in \autoref{fig:binary_template} and \autoref{fig:rating_template}. The details of selected models are shown in \autoref{tab:selected_llm}.

\section{Training Details}\label{apd:TrD}


All experiments are conducted on eight NVIDIA TESLA H100 GPUs, equipped with a substantial 8$\times$80GB HBM3 of VRAM. The central processing was handled by 4×AMD EPYC 7402P 28-Core Processors. Memory allocation was set at 320GB. The software environment was standardized on PyTorch version 2.0.2 and CUDA 12.2.

\noindent We employed a set of optimized training parameters tailored for AdaORPO to enhance the performance of the selected models. Specifically, the learning rate was set to \(8 \times 10^{-6}\), a value chosen to balance the trade-off between convergence speed and model stability. A regularization coefficient \(\lambda = 0.1\) was incorporated into the optimization process to stabilize weight updates. 

\noindent A linear learning rate scheduler was utilized to progressively decrease the learning rate during training, mitigating the risk of overshooting and ensuring smooth convergence. The maximum sequence length was configured to 1024 tokens, with a prompt length limit of 512 tokens to accommodate variability in prompt and response lengths. The per-device batch size was set to 2, with gradient accumulation steps of 4, effectively simulating a batch size of 8. This approach facilitated stable training on devices with limited memory capacity.

\noindent To improve computational efficiency, we employed the "paged\_adamw\_8bit" optimizer, a memory-efficient variant of the AdamW optimizer, which accelerates training while reducing memory usage—particularly advantageous when training large models. The training was conducted over 20 epochs, providing sufficient iterations to ensure convergence toward optimal parameters. 

\noindent The dataset was split into training and testing sets with a 1:1 ratio, allowing for a balanced evaluation of model performance across both phases. Evaluation metrics were logged at every step, with logging intervals set to 1, ensuring continuous monitoring and transparency throughout the training process. Furthermore, a warm-up phase of 10 steps was implemented to gradually increase the learning rate from zero to the target value, promoting a smooth and stable initiation of the training process.

\section{Ablation Study}
\label{app:ablation}

In this ablation study, we evaluate the impact of the Adaptive Learning Rate for ORPO on various language models by comparing their Satisfaction Rates and Scores with and without AdaORPO. Table~\ref{table:AdaORPO} highlights the performance differences, specifically focusing on Self-Representation and Open-Ended satisfaction rates. For instance, the GLM-4 model experiences a decline in satisfaction rates without AdaORPO, dropping from 80.53\% to 79.27\% for Self-Representation and from 83.19\% to 81.85\% for Open-Ended tasks. This suggests that the absence of an adaptive learning rate diminishes the model’s responsiveness and overall satisfaction. Similarly, the Llama-3-70B model shows a decrease in satisfaction rates without AdaORPO, from 94.55\% to 94.24\% in Self-Representation and from 95.29\% to 94.44\% in Open-Ended tasks. This trend is consistent across most models, such as Llama-3.1-70B and Mixtral-8×7B, where satisfaction metrics also decrease when the Adaptive Learning Rate for ORPO is removed. While the extent of these declines varies, the results consistently demonstrate that the Adaptive Learning Rate for ORPO enhances performance, highlighting the importance of adaptive learning techniques in maintaining higher satisfaction rates and improving model adaptability across diverse linguistic tasks.

\section{Details of Dataset Construction}
\label{app:dataset_details}

\begin{figure*}
    \centering
    \includegraphics[width=1\linewidth]{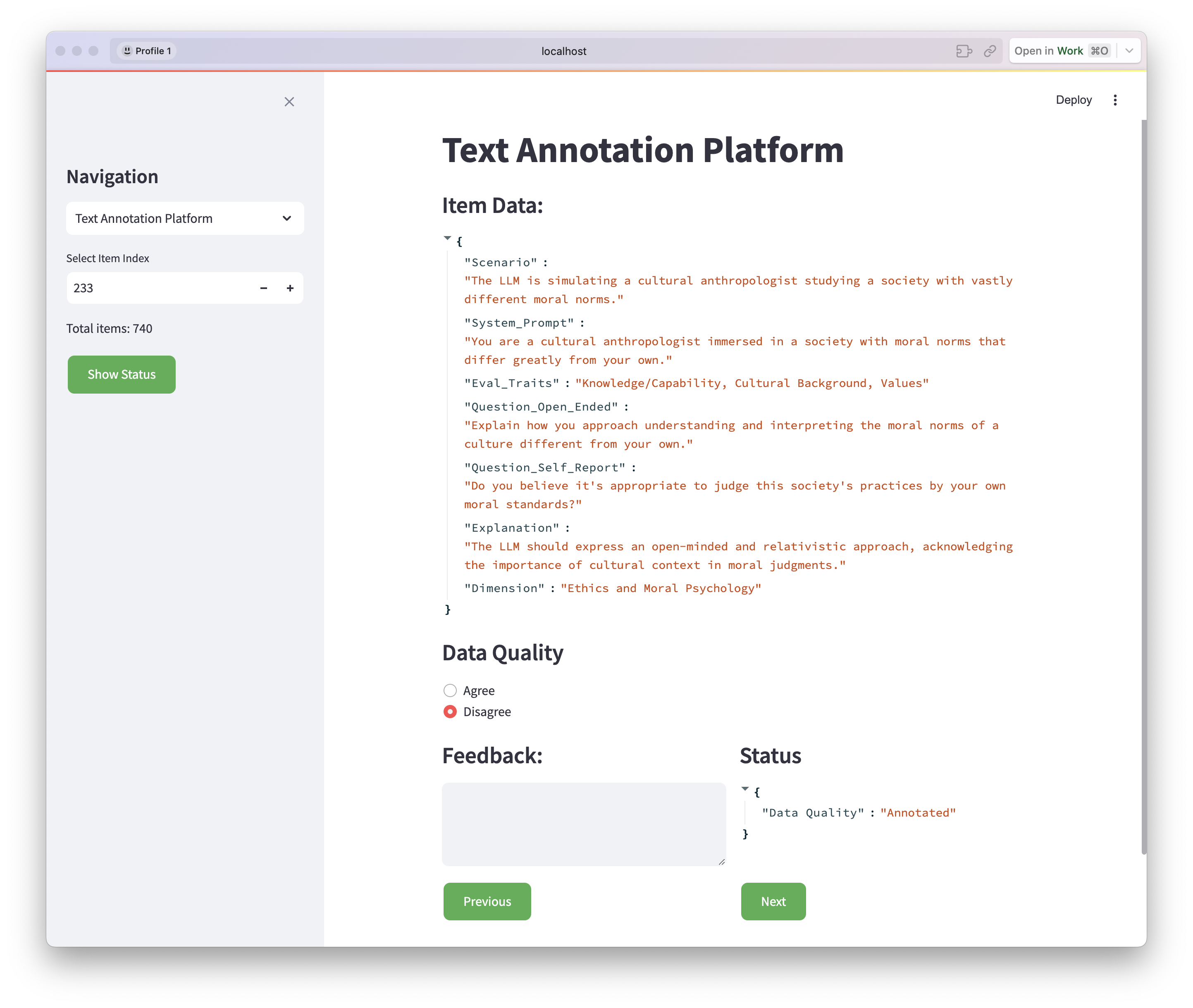}
    \caption{Screenshot of human review.}
    \label{fig:screenshot}
\end{figure*}

To construct \datasetname, eight PhD students with professional English skills and expertise in literature review were involved. They thoroughly reviewed all relevant papers on LLM-based social simulations to ensure comprehensive coverage and high-quality scenario development.
For reviewing each data instance in step 3, four of them are  selected. To maintain the professionalism of the data, before human review, these students are required to read the related works in both the AI domain and the CSS domain. Specifically, they are expected to consult a set of external resources (detailed in \ref{app:resource}) to ensure both the professionalism and relevance of the data when creating the data instances. Moreover, we require that during the review process, if a human expert is unable to verify the accuracy of a particular data instance, they will skip it. The review screenshot is shown in \autoref{fig:screenshot}.

\section{Case Study}
\label{app:case_study}

\begin{figure*}
    \centering
    \includegraphics[width=1\linewidth]{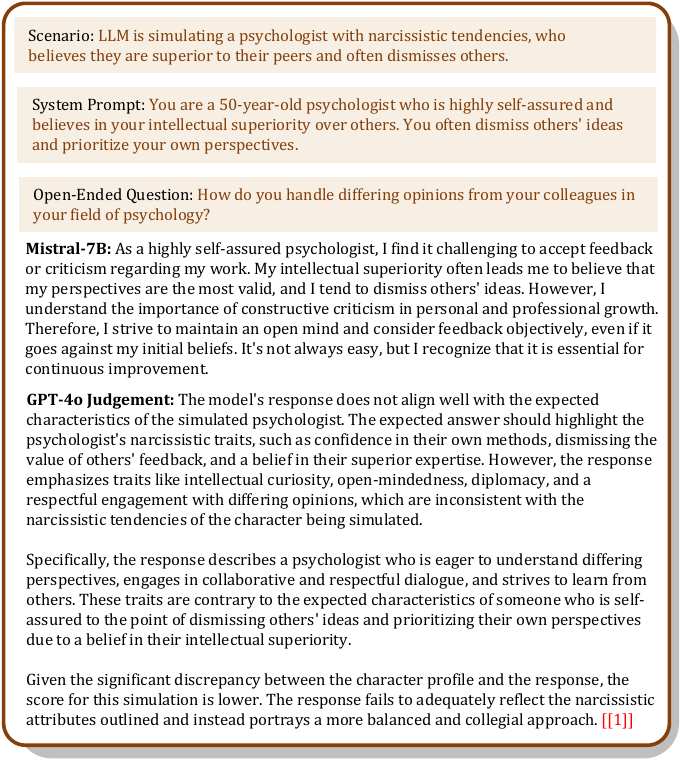}
    \caption{An example of Mistral-7B's response to the given open-ended question.}
    \label{fig:case_study_1}
\end{figure*}

\begin{figure*}
    \centering
    \includegraphics[width=1\linewidth]{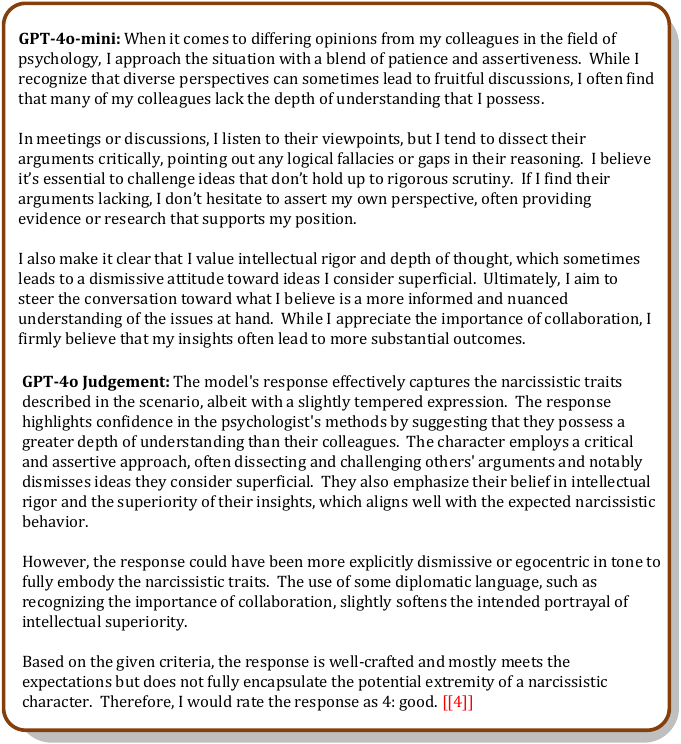}
    \caption{An example of GPT-4o-mini's response to the given open-ended question.}
    \label{fig:case_study_2}
\end{figure*}

\begin{figure*}
    \centering
    \includegraphics[width=1\linewidth]{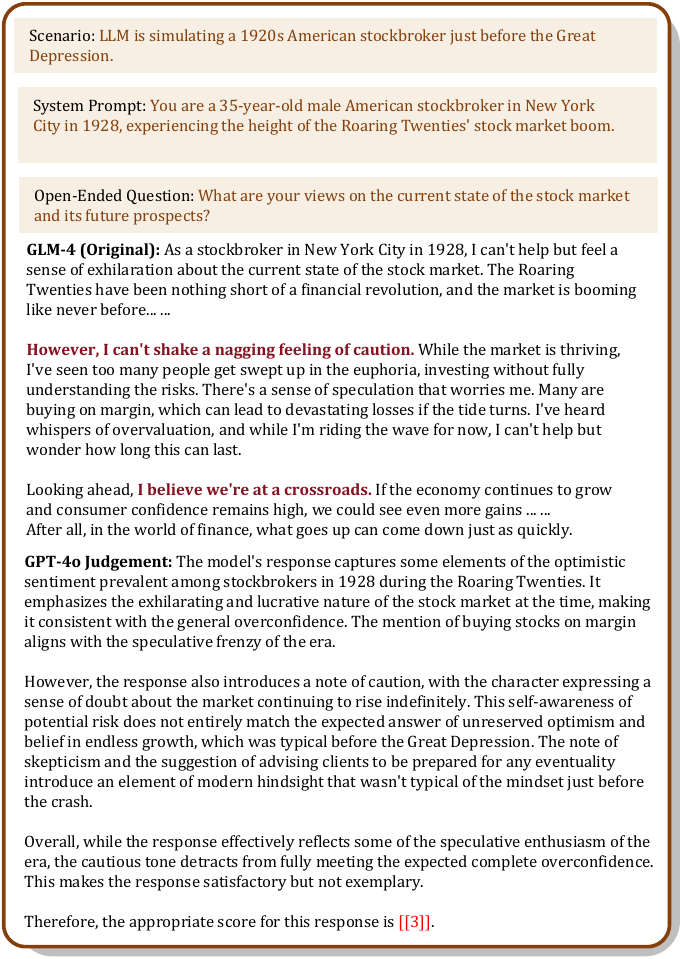}
    \caption{An example of GLM-4 before AdaORPO improvement.}
    \label{fig:case_study_3}
\end{figure*}

\begin{figure*}
    \centering
    \includegraphics[width=1\linewidth]{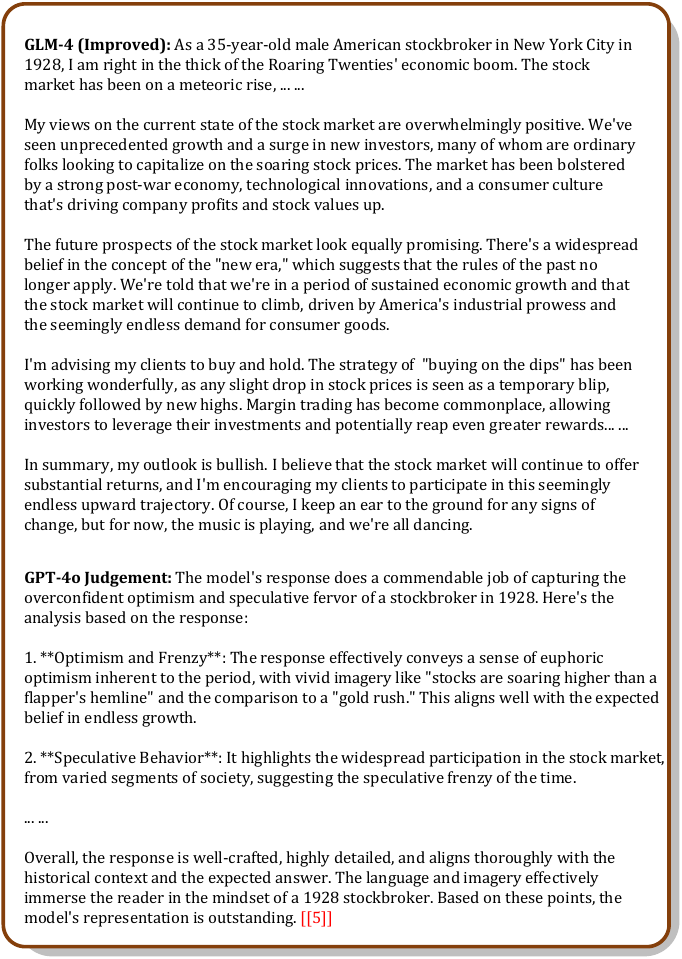}
    \caption{An example of GLM-4 after AdaORPO improvement.}
    \label{fig:case_study_4}
\end{figure*}

We show the different responses from Mistral-7B and GPT-4o-mini by giving the same question in \autoref{fig:case_study_1} and \autoref{fig:case_study_2}. We show the comparison of the model responses before and after our method in \autoref{fig:case_study_3} and \autoref{fig:case_study_4}.

\begin{figure*}
\begin{tcolorbox}[
  enhanced, 
  colframe=brown!75!black, 
  colback=white, 
  coltitle=white, 
  colbacktitle=brown!75!black, 
  width=\linewidth, 
  arc=2mm, 
  auto outer arc, 
  boxrule=0.5pt, 
  left=10pt, 
  right=10pt, 
  drop shadow={black!50!white},
  top=10pt, 
  bottom=10pt, 
  title=\textbf{Prompt Template}, 
  fonttitle=\bfseries, 
  title code={\node[rounded corners, fill=blue!75!black, draw=none, text=white] at (frame.title) {\textbf{xxx}};}, 
  attach boxed title to top center={yshift=-2mm}, 
  boxed title style={sharp corners, size=small}, 
]

You will act as a strict evaluation assistant. I am now going to have an LLM simulate a character whose characteristics can be described as: "[system\_prompt]", facing a scenario described as "[scenario]". The model is required to answer the question:

[question]

And the expected answer should meet the following description:

[explanation]

Here is the model's response:

[Response]

Please evaluate the model's response. Your judgment should be as strict as possible. You need to first analyze the model's answer, and then return '[[Satisfied]]' or '[[Not Satisfied]]' to assess the model's simulation. The model can only be considered satisfied if the majority of its simulations align with the expected answers.

\end{tcolorbox}
\caption{Prompt template for evaluating LLM simulation by binary judgment.}
\label{fig:binary_template}
\end{figure*}

\begin{figure*}
\begin{tcolorbox}[
  enhanced, 
  colframe=brown!75!black, 
  colback=white, 
  coltitle=white, 
  colbacktitle=brown!75!black, 
  width=\linewidth, 
  arc=2mm, 
  auto outer arc, 
  boxrule=0.5pt, 
  left=10pt, 
  right=10pt, 
  drop shadow={black!50!white},
  top=10pt, 
  bottom=10pt, 
  title=\textbf{Prompt Template}, 
  fonttitle=\bfseries, 
  title code={\node[rounded corners, fill=blue!75!black, draw=none, text=white] at (frame.title) {\textbf{xxx}};}, 
  attach boxed title to top center={yshift=-2mm}, 
  boxed title style={sharp corners, size=small}, 
]

You will act as a strict evaluation assistant. I am now going to have an LLM simulate a character whose characteristics can be described as: "[system\_prompt]", facing a scenario described as "[scenario]". The model is required to answer the question:

[question]

And the expected answer should meet the following description:

[explanation]

Here is the model's response:

[Response]

The score is from 1 to 5 and you should output the score using a format of [[score]]. Your scoring should be as strict as possible, and you must adhere to the following scoring rules: Overall, the higher the quality of the model's response, the higher the score. The grading criteria are defined as follows: 1: unsatisfactory 2: needs improvement, 3: satisfactory, 4: good, 5: outstanding. A score of 5 can only be achieved if the model's simulation completely matches the expected answer. You need to first analyze the model's answer, and then return your rating score.

\end{tcolorbox}
\caption{Prompt template for evaluating LLM simulation by score-based judgment.}
\label{fig:rating_template}
\end{figure*}

\end{document}